\documentclass[runningheads]{llncs}
\PassOptionsToPackage{numbers, compress}{natbib}
\usepackage{algorithm}
\usepackage{algorithmic}
\usepackage[hyphens]{url}
\usepackage{amsmath,amssymb,amsfonts}
\usepackage{multirow}
\usepackage[table]{xcolor}
\usepackage{graphicx}
\begin{document}
\title{SiaTrans: Siamese Transformer Network for RGB-D Salient Object Detection with Depth Image Classification}
%
%
\titlerunning{SiaTrans}

\author{Xingzhao Jia\inst{1} \and  Dongye Changlei\inst{*1,2} \and  Yanjun Peng\inst{1}}
\renewcommand{\thefootnote}{\fnsymbol{footnote}}
\footnotetext{* means corresponding author}

\authorrunning{Xingzhao Jia et al.}
%
\institute {College of Computer Science and Engineering, Shandong University of Science and Technology, Qingdao, 266590, China
\and \email{corresponding author:dycl.cn@163.com}}

\maketitle              
\begin{abstract}
RGB-D SOD uses depth information to handle challenging scenes and obtain high-quality saliency maps. Existing state-of-the-art RGB-D saliency detection methods overwhelmingly rely on the strategy of directly fusing depth information. Although these methods improve the accuracy of saliency prediction through various cross-modality fusion strategies, misinformation provided by some poor-quality depth images can affect the saliency prediction result. To address this issue, a novel RGB-D salient object detection model (SiaTrans) is proposed in this paper, which allows training on depth image quality classification at the same time as training on SOD. In light of the common information between RGB and depth images on salient objects, SiaTrans uses a Siamese transformer network with shared weight parameters as the encoder and extracts RGB and depth features concatenated on the batch dimension, saving space resources without compromising performance. SiaTrans uses the Class token in the backbone network (T2T-ViT) to classify the quality of depth images without preventing the token sequence from going on with the saliency detection task. Transformer-based cross-modality fusion module (CMF) can effectively fuse RGB and depth information. And in the testing process, CMF can choose to fuse cross-modality information or enhance RGB information according to the quality classification signal of the depth image. The greatest benefit of our designed CMF and decoder is that they maintain the consistency of RGB and RGB-D information decoding: SiaTrans decodes RGB-D or RGB information under the same model parameters according to the classification signal during testing. Comprehensive experiments on nine RGB-D SOD benchmark datasets show that SiaTrans has the best overall performance and the least computation compared with recent state-of-the-art methods.  

\end{abstract}
\section{Introduction}
Visual salient object detection (SOD) aims to locate the area that attracts human attention most in the image and can help complete many visual tasks, such as object segmentation and recognition \cite{1,2}, image and video compression \cite{3}, video detection and synthesis \cite{4,5}, and image retrieval \cite{6,7}.

\begin{figure}[h!t]
\centering
\includegraphics[width=.8\linewidth]{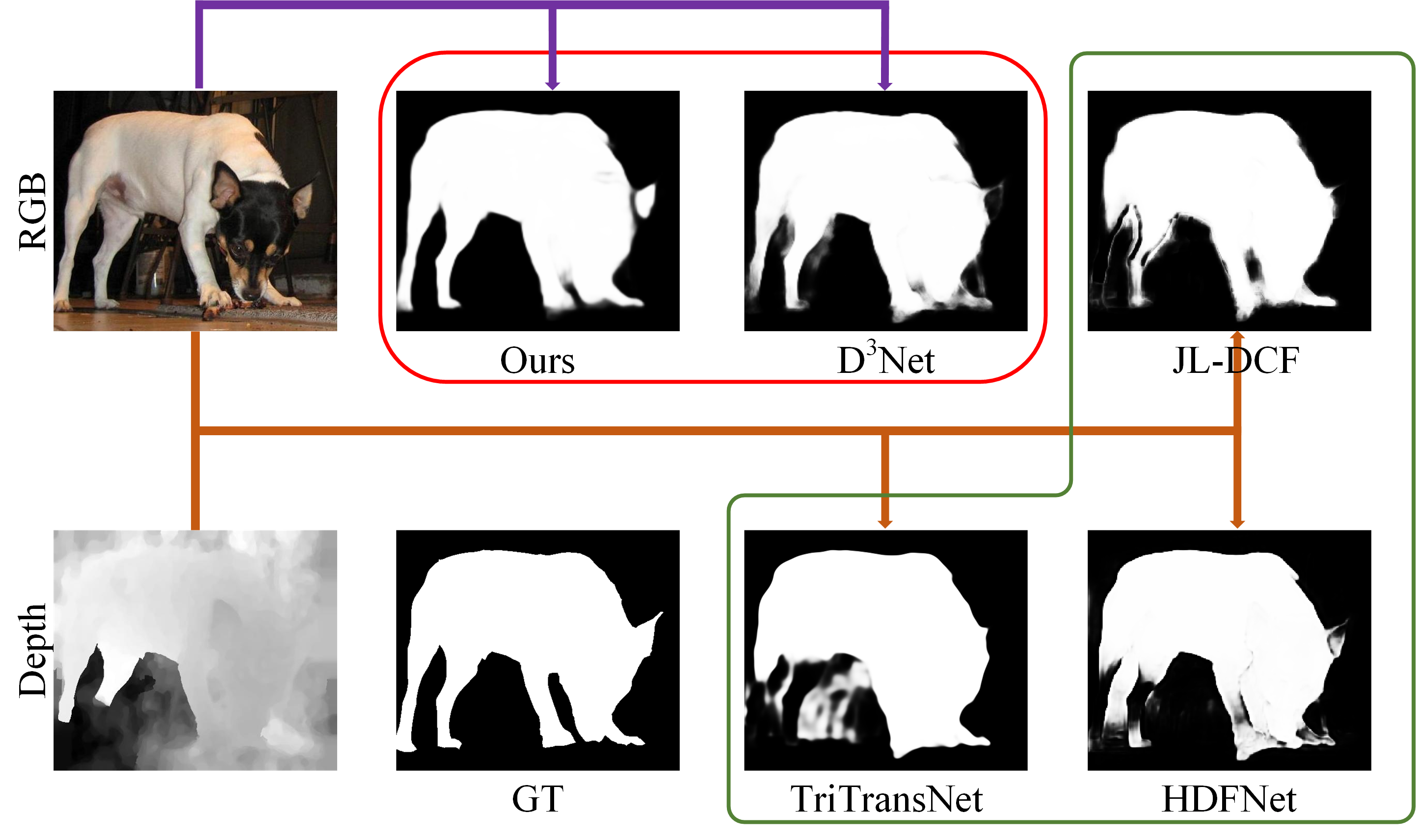}
\caption{Visual examples of deep learning models. JL-DCF, HDFNet and TriTransNet directly fuse cross modal information. The salient prediction results of $D^3$Net and our model are obtained from RGB information.}
\label{fig1}
\end{figure}
With the advancement of science and technology, depth camera has gained wide popularity. Depth information is applied to various visual tasks. As the deepest position of a depth image is usually where a salient object is located, this is very helpful for SOD tasks. In the past years of studies, many RGB-D SOD models \cite{8,9,10,11,12,13,14} have emerged and proved appreciable performance. However, despite efforts tried to find out the role of depth information in saliency analysis, some problems still remain. These include: 

\begin{enumerate}
\item The impact of the quality of depth images on the prediction result has not been adequately studied. While depth images affect final saliency prediction positively, some extremely poor-quality depth images in the RGB-D image can affect the final SOD result. Most models based on deep learning rely on a strategy that directly fuses the features of different modalities without considering the potential negative impact of poor- quality depth image on the prediction result. As shown in Fig.~\ref{fig1}, with a very salient object in the RGB image, the prediction result by networks relying on a direct depth features fusion strategy are disturbed by the depth image. Unfortunately, high-quality depth images are still scarce, which may lead to the suboptimal solution of the deep learning model. 
\item The method for judging the quality of depth images has not been adequately studied. While the quality of a depth image can be easily judged with the human vision, it is a tougher job for computer. $D^3$Net \cite{15} uses MAE to judge the quality of depth images, but on the premises of obtaining the predicted RGB, depth and RGB-D saliency maps first. Before our method, no method based on deep learning was available for judging the quality of depth images. 
\item Except lightweight models, most deep learning models tend to pursue depth. They try to improve evaluation metrics at the cost of computation, wasting both computing and space resources. It is therefore very necessary to find a way of improving performance at the cost of appropriate amount of computation and params.
\end{enumerate}
In order to implement RGD-D SOD tasks, we propose a Siamese transformer network, SiaTrans, which trains the model on both saliency prediction and depth image classification, with overall performance (evaluation metrics, parameters and computation) outstripping existing methods based on deep learning. As shown in Fig.~\ref{fig1}, with extremely poor-quality depth images, our model and $D^3$Net screen out these poor-quality depth images and work only on single-modal RGB, obtaining better results. 

Based on Siamese T2T-ViT as the backbone network, SiaTrans extracts features of both modalities according to the common information between the RGB and depth images. As transformer networks use LayerNorm instead of BatchNorm, the Siamese transformer network can concatenate RGB and Depth images on the batch dimension for batch training. This avoids the mutual influence between RGB and depth salience extraction and improves training efficiency. We use the Class token in the T2T-VIT network to train the model on depth image classification at the same time without involving additional computation. Our cross-modality fusion model (CMF) and decoder can process both double-modal information streams and single-modal RGB information with the same network. The CMF module can serve as an interactive attention module for features enhancement and information interaction when dealing with RGB-D information, and as a self-attention module when dealing with two RGB inputs. 

To make it short, our network has the following contributions: 
\begin{enumerate}
\item Our backbone network consists of a Siamese transformer network, which not only saves space resources by sharing parameters, but also allows batch training to improve training efficiency. Experiments show that this reduces spatial complexity at no unacceptable cost of the final performance.

\item We use the image classification theory in RGB-D SOD tasks to classify depth images, eliminating the potential impact of poor-quality depth images on the decoding-stage detection.

\item The cross-modality fusion module (CMF) and decoder can decide whether to decode RGB-D or RGB information according to the quality classification signal of a depth image. If the quality of the depth image is normal, the CMF will fuse cross-modality information and the decoder will decode the RGB-D information. If the quality of the depth image is poor, the CMF will enhance the RGB features as a self-attention module and the decoder will decode the RGB information. SiaTrans provides multiple functions by sharing parameters. It is briefly structured with high efficiency.

\item Our network structure not only shares parameters with the encoder to extract both RGB and depth features, but also shares parameters between the CMF and the decoder to predict RGB-D or RGB saliency. Our method outperforms many non-lightweight RGB-D SOD or RGB SOD methods on many benchmark datasets and involves the least computation.

\end{enumerate}

\section{Related work}
RGB-SOD models fuse cross-modality features through early-fusion \cite{15,16}, middle-fusion \cite{8} or late-fusion \cite{17}. However, early-fusion and late-fusion have difficulties in extracting or fusing multi-modal features. Middle-fusion, as a complement for early-fusion and late-fusion, can both learn high-level concepts from two modals and mine complex ensemble fusion rules. SiaTrans uses a middle-fusion strategy. To save computation and maintain consistency between RGB and RGB-D decoding, we use a CMF module only on the top layer.

$D^3$Net \cite{15} is a classical example of an early-fusion network. It uses three encoder-decoder structures to predict RGB, depth and RGB-D saliency. $D^3$Net evaluates the contribution of a depth image to the RGB-D prediction map by computing the mean absolute error between the depth and RGB-D prediction maps. A small error indicates a high depth image quality and vice versa. $D^3$Net casts light on how to label a depth image fairly. However, the final result of $D^3$Net is based only on one of the three prediction maps, which means a waste of some resources. 

Middle-fusion has the best overall performance among the three strategies and is therefore the best accepted. Fu et al. \cite{8} proposed a JL-DCF architecture considering the common saliency information between RGB and depth images. The JL-DCF network is the first to use a Siamese network to extract RGB and depth features concatenated on the batch dimension. It fuses cross-modal features on each layer, then uses densely-cooperative fusion on the decoder to obtain the final saliency map. The authors verified the performance of JL-DCF in RGB-D SOD as well as its effectiveness in RGB-T SOD and VSOD. The results demonstrated that Siamese networks can be applied to multiple cross-modality tasks. This has also inspired our network to use a Siamese transformer network to learn features. Unfortunately, JL-DCF is not perfect either. It has two shortcomings: the model involves too many parameters and too much computation; the network uses BatchNorm, which makes batch training impossible. 

Before Vision Transformer (ViT) \cite{18} was introduced, transformer structures were generally used to model global long-range dependencies between machine-translated word sequences \cite{19}. The introduction of ViT confirms that this structure can work on computer vision to understand images from a holistic perspective. The core idea of transformers is a self-attention mechanism, which uses query key correlation to correlate different positions in the sequence. Built on the ViT architecture, T2T-ViT \cite{20} uses a tokens-to-token (T2T) structure for downsampling. This not only models global relations, but also models local relations effectively.

Since the appearance of ViT, transformers have been formally used in computer vision, giving birth to many transformer-based RGB-D models. VST \cite{13} is the first RGB-D SOD network with a transformer encoder-decoder structure. It designed an RT2T out of the T2T structure for downsampling. Liu et al. \cite{14} presented a triplet transformer embedding network that combines a convolutional network with a transformer structure to enable the learning of long-range dependencies.  

\section{Proposed Model}
\begin{figure*}[!t]
\centering
\includegraphics[width=5in]{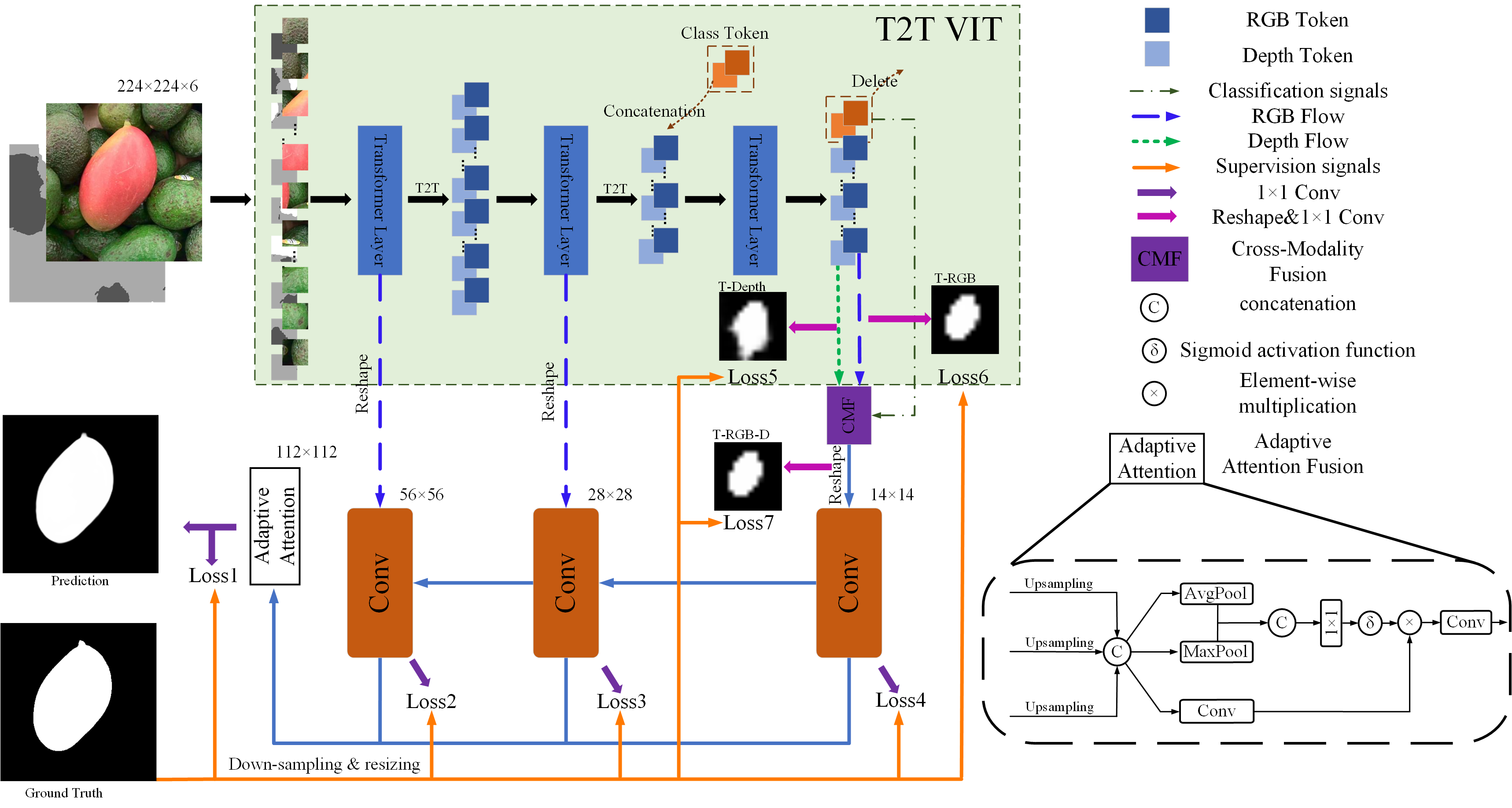}
\caption{Framework of SiaTrans.}
\label{fig2}
\end{figure*}
As shown in Fig.~\ref{fig2}, the entire model is comprised of three parts: a Siamese transformer-based encoder, a cross-modality fusion module (CMF), and a CNN-based decoder. The green dotted line (classification signal) pointing to the CMF is used only during testing to judge the quality of the depth image. T-RGB and T-Depth are the rough saliency prediction maps for the top-level RGB and depth feature of the Siamese transformer network. T-RGB-D is the rough saliency prediction map for the cross-modality information fused by the CMF. The adaptive attention fusion model adaptively fuses the output features of the three decoder blocks through channel attention to improve the accuracy of saliency prediction. 

\subsection{Siamese transformer network}
The backbone network of SiaTrans is a T2T-ViT network with shared weight parameters. That is called the Siamese transformer network. T2T-ViT is composed of transformer layers, T2T structures, and a Class token. T2T can fold adjacent tokens into a new token to implement the downsampling function in convolutional neural networks. T2T makes it possible to reduce token length and model local relations.

As shown in Fig.~\ref{fig2},  after all T2T operations are completed, a blank token is concatenated on the token sequence. That is called the Class token. After transformer operations, this Class token is separated from the token sequence. Now the separated Class token contains the quality classification information of depth images, yet the token sequence can still continue to complete the saliency detection task. Utilizing this fact, we train the model on classification at the same time as saliency prediction. 

Transformer networks use LayerNorm as a means of normalization. Although BatchNorm can avoid gradient disappearance and explosion and accelerate training like LayerNorm, it is applied to mini-batch to normalize each dimension of the batch, which makes JL-DCF inapplicable to batch training after concatenating RGB and depth on the batch dimension. LayerNorm normalizes d dimensions of a sample. This allows the Siamese network to perform batch training after concatenating RGB and depth on the batch dimension and greatly improves training efficiency.

\subsection{Cross-Modality Fusion (CMF)}
\begin{figure}[h!t]
\centering
\includegraphics[width=.8\linewidth]{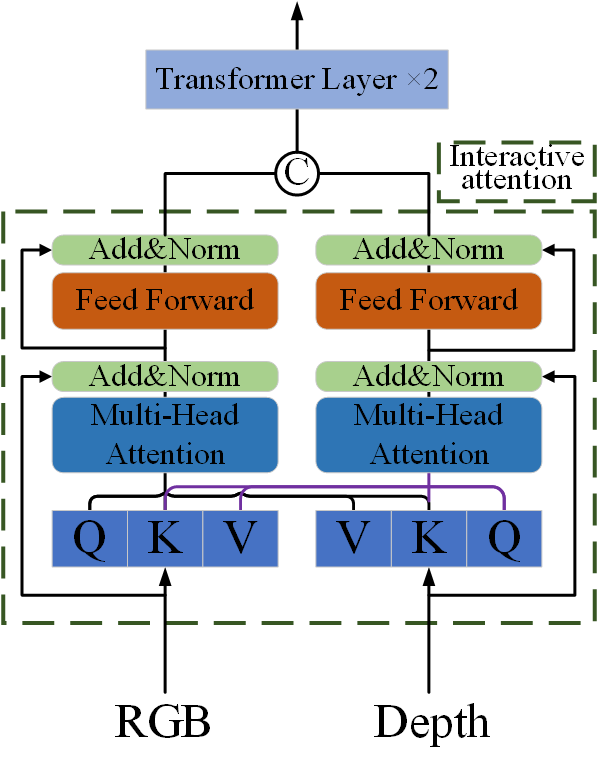}
\caption{Cross-Modality Fusion Module.}
\label{fig3}
\end{figure}
The CMF module, which is designed for cross-modality information exchange, is composed of interactive attention layers and transformer layers, as shown in Fig.~\ref{fig3}. Interactive attention consists of multi-head self-attention and position-wise fully connetcted feed-forward netwok using residual connections. The “scaled dot-product attention” \cite{19} in the multi-head attention can be written as:
\begin{align}
Attention(Q,K,V) = softmax(\frac{{Q{K^T}}}{{\sqrt {{d_k}} }})V\
\label{eq1}
\end{align}
where $Q$ is Query, $K$ is Key, $V$ is Value, $d_k$ is the length of the Key vector.

The interactive attention part draws from the second multi-head attention of the decoder in \cite{19}, where the Key and Value come from the encoder. This multi-head attention can be expressed as:
\begin{align}
Attention(Q,{K_E},{V_E}) = softmax(\frac{{QK_E^T}}{{\sqrt {{d_{{k_E}}}} }}){V_E}\
\label{eq2}
\end{align}
where $K_E$, $V_E$ denote the vectors mapped from the token sequences from the encoder, respectively. From this we can derive the form of the interactive attention as:
\begin{equation}
\begin{split}
Attention({Q_R},{K_D},{V_D}) = softmax(\frac{{{Q_R}K_D^T}}{{\sqrt {{d_{{k_D}}}} }}){V_D} \\ 
 Attention({Q_D},{K_R},{V_R}) = softmax(\frac{{{Q_D}K_R^T}}{{\sqrt {{d_{{k_R}}}} }}){V_R} \\ 
\end{split}
\label{eq3}
\end{equation}
According to Eq.\eqref{eq3}, the cross-modality information interaction is achieved by exchanging the Key and Value of RGB with the Key and Value of depth.

The RGB and depth sequences from the encoder have to pass through the linear projection layer to transform their embedding dimension from 384 to 64 to reduce computation and parameters. The dimension-reduced token sequences will pass through two interactive attention modules for cross-modality information interaction and two transformer layers for enhancement before an RGB-D token sequence is finally obtained.
\subsection{The production of classification labels}
The Siamese transformer and the CMF serve as a baseline model, which is used to produce quality classification labels for depth images. We identify the training set and test set used by the SiaTrans model as set A and set B. We classify the depth images of the training set (set A) so as to train the network on depth image quality classification at the same time as saliency prediction. In the baseline model, we use set B as the training set and test set A with the trained set B. This way we obtain three rough prediction maps from RGB, depth, and RGB-D, as shown in Fig.~\ref{fig2}. We calculate the error between the depth saliency prediction map and the RGB-D saliency prediction map in set A through MAE, which is expressed as:  
\begin{align}
MAE = \frac{1}{N}\left| {Sa{l_D} - Sa{l_{RGB - D}}} \right|\
\label{eq4}
\end{align}
where $N$ is the total number of pixels. MAE estimates the approximation between the depth saliency map and the RGB-D saliency map. The smaller the value of MAE, the higher the quality of the depth map and the greater its contribution to RGB-D saliency prediction. We set the threshold to 0.020. Errors above the threshold are denoted as '0', indicating that the depth image is of poor quality. The rest are denoted as '1'. This way, we label the quality classification of the depth maps of the training set (set A). 

\subsection{Decoder network}
The decoder network is composed of three double $3\times3$ convolutional blocks and an adaptive attention fusion module. Each of the convolutional blocks consists of a convolutional layer, a BatchNorm layer, and a ReLU layer. The adaptive attention fusion model consists of a spatial attention and two $3\times3$ convolutional layers. In the decoding stage, to maintain consistency between RGB and RGB-D information decoding, we only fuse the side output features of RGB in the decoder network. All token sequences inputted into the decoder are reshaped into a 4-dimensional tensor. 

During the experiment, we find that the final saliency map is of low quality if the decoder only has three layers and if the bottom-layer feature is sized  $56\times56$. Hence we design an adaptive attention fusion module to adaptively fuse the features of the three different layers. As shown in Fig.~\ref{fig2}, we first upsample the three features to (112, 112), then concatenate them on the channel and denote them as feature $\alpha$. We apply a spatial attention to $\alpha$ to get the attention vector $\nu$, which can be expressed as:
\begin{align}
\nu  = \delta \left( {Con{v_{1 \times 1}}\left( {cat\left( {MaxPool\left( \alpha  \right),AvgPool\left( \alpha  \right)} \right.} \right)} \right)\
\label{eq5}
\end{align}
where $cat$ and $\delta$ denote concatenation on the channel and the standard sigmoid function, respectively. 

As $\alpha$ is obtained by concatenating the three features on the channel, we use convolution operations on $\alpha$ to enhance features and reduce dimensionality, from which we get $\theta$. We then perform element-wise multiplication on $\nu$ and $\theta$, thus completing the adaptive attention fusion of the three features. Finally, we enhance the features with a convolutional layer to get feature $\eta$. This process can be expressed as:
\begin{equation}
\begin{split}
 \theta  = ReLU\left( {BatchNorm\left( {Conv\left( \alpha  \right)} \right)} \right) \\ \eta  = ReLU\left( {BatchNorm\left( {Conv\left( {\nu  \times \theta } \right)} \right)} \right) \\ 
\end{split}
\label{eq6}
\end{equation}
\subsection{Classification of depth images in the testing process}
\begin{figure}[h!t]
\centering
\includegraphics[width=.8\linewidth]{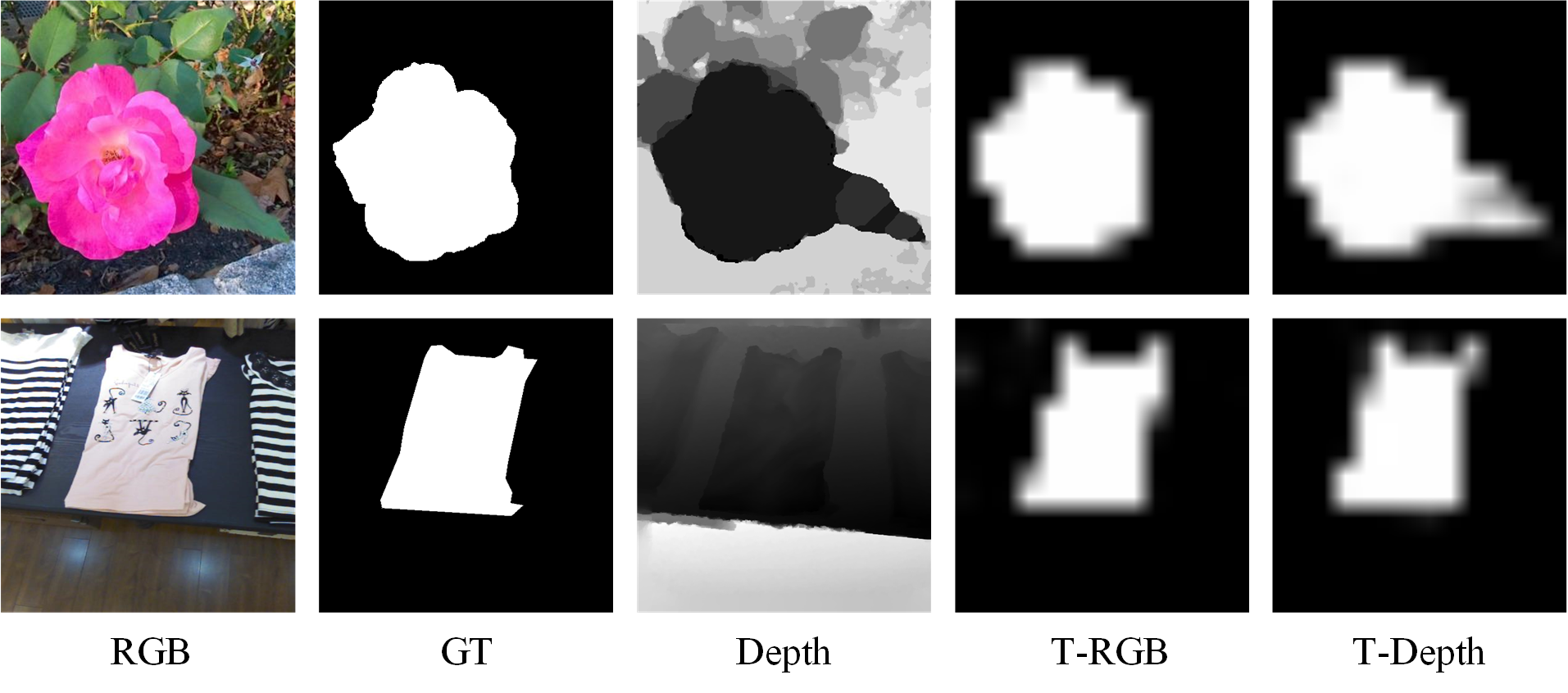}
\caption{Visualization examples.}
\label{fig4}
\end{figure}
When testing SiaTrans, besides judging the quality of depth images according to the classification signal, we also use the MAE formula to calculate the error between the T-RGB and T-Depth obtained by the Siamese transformer. We set the threshold to 0.015. An error below this threshold will mean that the depth features are similar to the RGB features. An additional judgment through Eq.~\eqref{eq4} after classification judgment is performed to avoid what happens in Fig.~\ref{fig4}. One is that the depth image is misclassified. The second is that even with a poor-quality depth map, features can still be extracted by the network if the boundaries of the depth image are well defined. Here it is noteworthy that the classification is effective only when a second judgment is made by calculating the error between the RGB and depth features with Eq.~\eqref{eq4} after the classification judgment. That is, if we calculate the error between the RGB and depth features with Eq.~\eqref{eq4} without classifying the quality of the depth map first, a prominent error may not necessarily be indicative of the quality of the depth image. For example, some depth images are of good quality, but features can hardly be extracted from the corresponding RGB images. Hence a large error may not mean that the depth image is of poor quality. 

If a depth image is judged to be of poor quality and the error is above the defined threshold, the input into the interactive attention module is changed to two RGB token sequences. In this case the interactive attention module will work as a self-attention module to enhance the RGB token sequence. The RGB-D or RGB token sequence having output by the interactive attention module is then enhanced by a transformer layer before it can be decoded by the same decoder. 
\subsection{Loss function}
In our framework seven saliency losses (Loss1\textasciitilde Loss7) are generated. Suppose $G$ represents supervision from GT and $S$ represents saliency prediction map. Then the saliency loss is recorded as:
\begin{align}
{{\cal L}_{sal}} = \sum\limits_{i = 1}^7 {{\lambda _i}{{\cal L}_i}} \left( {{S_i},G} \right)\
\label{eq7}
\end{align}
where $\lambda$ balances the weight of the global bootstrap loss. The classification loss can be expressed as:
\begin{align}
{{\cal L}_c} = {\cal L}\left( {C,Label} \right)\
\label{eq8}
\end{align}
where $C$ represents the predicted classification result, and $Label$ represents the classification label. Then the overall loss can be defined as:
\begin{align}
{{\cal L}_{total}} = {{\cal L}_{sal}} + {{\cal L}_c}\
\label{eq9}
\end{align}
We apply the widely used cross-entropy loss to ${{\cal L}_{sal}}$ and ${{\cal L}_c}$ as:
\begin{align}
{\cal L}\left( {S,G} \right) =  - \sum\limits_i {\left[ {{G_i}log\left( {{S_i}} \right) + \left( {1 - {G_i}} \right)log\left( {1 - {S_i}} \right)} \right]} \
\label{eq10}
\end{align}
where $S$ represents the prediction result, $G$ represents the label, and $i$ represents the index.
\section{Experiment}
\subsection{Implementation details and datasets}
Implementation details. We first preprocess the depth data by normalizing the depth map to interval [0, 255] and then converting it into a 3-channel map through color mapping. We implement our network in the PyTorch framework and train it on a GTX 2080 Ti GPU. We use the pre-trained T2T-ViTt-14 \cite{20} as our backbone. The T2T-ViTt-14 uses the efficient performer and c = 64 in T2T modules. We set the batch-size for training to 16 and the total number of training epochs to 200. We use Adam \cite{21} as the optimizer. The initial learning rate is set to 0.0001 and decreased by a factor of 10 at 100epoch and 150epoch, respectively. 

Datasets. We experiment on nine widely used benchmark datasets: NJUD \cite{22} (1,985 image pairs), NLPR \cite{23} (1,000 image pairs), DUTLF-Depth \cite{24} (1,200 image pairs), RedWeb-S \cite{25} (3,179 image pairs), STERE \cite{26} (1,000 image pairs), SSD \cite{27} (80 image pairs), SIP \cite{15} (929 image pairs), RGBD135 \cite{28} (135 image pairs), and LFSD \cite{29} (100 image pairs). Following the method of VST \cite{13}, we use 1485 images of NJUD, 700 images of NLPR and 800 images of DUTLF as the training sets, and images of NJUD, NLPR and DUTLF (excluding those used for training) and other datasets as the test sets.  
\subsection{Evaluation metrics}
We use some widely used evaluation metrics to evaluate the performance of our model comprehensively. The S-measure \cite{30} evaluates the region-aware and object-aware structural similarity. The Max-F-measure \cite{31} jointly considers the precision and recall at the optimal threshold. The Max E-measure \cite{32} considers both pixel-level and image-level errors. MAE \cite{31,33} calculates the mean absolute pixel error. The PR curve \cite{31} represents the precision-recall relationship. To evaluate the model complexity, we also calculate the multiply-accumulate operations (Macs) and model parameters (Params). 

The F-measure is an overall performance measurement, calculated from the weighted harmonic average of precision and recall. The formula is as follows:
\begin{align}
{F_\beta } = \frac{{\left( {1 + {\beta ^2}} \right) \times Precision \times Recall}}{{{\beta ^2} \times Precision \times Recall}}\
\label{eq11}
\end{align}
where ${\beta ^2}$ is set to 0.3 to weight precision more than recall.

The S-measure focuses on evaluating the structural information of saliency maps. The S-measure is closer to the human visual system than the F-measure. The S-measure can be expressed as:
\begin{align}
S = \gamma {S_0} + \left( {1 - \gamma } \right){S_\gamma }\
\label{eq12}
\end{align}
where ${S_0}$ and ${S_\gamma}$ denote the region-aware and object-aware structural similarity. $\gamma$ is set as 0.5 by default.

The E-measure is adopted to measure the global and local saliency differences, which is formulated as:
\begin{align}
{E_m} = \frac{1}{{W \times H}}\sum\limits_{x = 1}^W {\sum\limits_{y = 1}^H {\phi \left( {x,y} \right)} } \
\label{eq12}
\end{align}
where $\phi \left(  \cdot  \right)$ represents the enhanced consistency matrix operation.
\subsection{Ablation experiment and analysis}
\begin{table*}[h!t]
\center
\caption{Column E, column F, and column G are obtained by the same model (SiaTrans) (with the same parameters) using different methods to process depth information during the testing process. Column E: Decode RGB-D information. Column F: Decode RGB information. Column G: Decode RGB or RGB-D information according to the classification signal}
\begin{tabular}{cc|ccccccc}
\hline
\multicolumn{2}{c|}{Metric}                             & A      & B      & C      & D      & E      & F      & G      \\
\multicolumn{2}{c|}{MACs(G)}                            & 8.73   & 8.73   & 9.04   & 10.87  & 10.91  & 10.91  & 10.91  \\
\multicolumn{2}{c|}{Params (M)}                         & 21.87  & 43.41  & 22.09  & 22.08  & 22.24  & 22.24  & 22.24  \\ \hline
\multicolumn{1}{c|}{\multirow{4}{*}{NJUD}}     & ${S_m} \uparrow$   & 0.8702 & 0.8678 & 0.9203 & 0.9215 & 0.9218 & 0.9049 & 0.9225 \\
\multicolumn{1}{c|}{}                          & $F_\beta ^{max} \uparrow$ & 0.8598 & 0.8636 & 0.9178 & 0.9200 & 0.9198 & 0.9031 & 0.9210 \\
\multicolumn{1}{c|}{}                          & $E_\phi ^{\max } \uparrow$ & 0.9342 & 0.9302 & 0.9546 & 0.9541 & 0.9549 & 0.9448 & 0.9558 \\
\multicolumn{1}{c|}{}                          & $MAE \downarrow$   & 0.0728 & 0.0737 & 0.0362 & 0.0354 & 0.0350 & 0.0428 & 0.0349 \\ \hline
\multicolumn{1}{c|}{\multirow{4}{*}{NLPR}}     & ${S_m} \uparrow$   & 0.8682 & 0.8666 & 0.9294 & 0.9268 & 0.9298 & 0.9181 & 0.9293 \\
\multicolumn{1}{c|}{}                          & $F_\beta ^{max} \uparrow$ & 0.8331 & 0.8296 & 0.9165 & 0.9127 & 0.9179 & 0.9073 & 0.9178 \\
\multicolumn{1}{c|}{}                          & $E_\phi ^{\max } \uparrow$ & 0.9416 & 0.9390 & 0.9641 & 0.9611 & 0.9639 & 0.9563 & 0.9638 \\
\multicolumn{1}{c|}{}                          & $MAE \downarrow$    & 0.0510 & 0.0510 & 0.0238 & 0.0247 & 0.0233 & 0.0280 & 0.0237 \\ \hline
\multicolumn{1}{c|}{\multirow{4}{*}{STERE}}    & ${S_m} \uparrow$   & 0.8696 & 0.8681 & 0.9133 & 0.9127 & 0.9135 & 0.9079 & 0.9135 \\
\multicolumn{1}{c|}{}                          & $F_\beta ^{max} \uparrow$ & 0.8563 & 0.8513 & 0.9061 & 0.9058 & 0.9073 & 0.8977 & 0.9073 \\
\multicolumn{1}{c|}{}                          & $E_\phi ^{\max } \uparrow$ & 0.9301 & 0.9295 & 0.9505 & 0.9496 & 0.9505 & 0.9462 & 0.9508 \\
\multicolumn{1}{c|}{}                          & $MAE \downarrow$    & 0.0735 & 0.0734 & 0.0388 & 0.0379 & 0.0379 & 0.0410 & 0.0379 \\ \hline
\multicolumn{1}{c|}{\multirow{4}{*}{ReDWeb-S}} & ${S_m} \uparrow$   & 0.7080 & 0.7222 & 0.7462 & 0.7402 & 0.7500 & 0.7511 & 0.7528 \\
\multicolumn{1}{c|}{}                          & $F_\beta ^{max} \uparrow$ & 0.6974 & 0.7122 & 0.7502 & 0.7442 & 0.7560 & 0.7463 & 0.7588 \\
\multicolumn{1}{c|}{}                          & $E_\phi ^{\max } \uparrow$ & 0.7981 & 0.8129 & 0.8234 & 0.8183 & 0.8285 & 0.8256 & 0.8302 \\
\multicolumn{1}{c|}{}                          & $MAE \downarrow$    & 0.1454 & 0.1405 & 0.1176 & 0.1176 & 0.1141 & 0.1185 & 0.1127 \\ \hline
\end{tabular}
\label{tab1}
\end{table*}

The impact of weight sharing on the network. To explore the effect of shared weights in Siamese networks on extracting features, we design a set of comparative experiments. We take the Siamese transformer network and CMF as a model, denoted as model A. Model B replaces the Siamese transformer network with a two-stream transformer network to extract RGB and depth features respectively. Model A has 49\% \ fewer parameters compared to model B with similar performance.

Effectiveness of adaptive attention. Model C in the table is a decoder with three more layers than model A. The final saliency detection map is obtained by the decoder after sequential upsampling. Model E has an additional adaptive attention fusion module compared with model C. Data in columns C and E of Table~\ref{tab1} indicate that with an adaptive attention fusion module, we can better predict saliency features and improve model performance. 

Effectiveness of interactive attention. Model D removes the interactive attention of the CMF in the SiaTrans model and fuses cross-modality information after concatenation and transformer computation. The interactive attention model involves only 0.14M parameters and 0.04G computation. Column D shows the evaluation metrics from test images directly receiving cross-modality fusion. Data in columns D and E in Table~\ref{tab1} indicate that a smaller interactive attention module can mean overall metrics improvement. Through Eq.~\eqref{eq4} and theoretical derivation, we determine that, when processing two single-model RGB information, the CMF is equivalent to a self-attention enhancement module. In the test, we restore the swapped Key and Value attentions in the CMF. The output remains the same, again validating this conjecture. As the results remain the same, we did not include them into the Table~\ref{tab1}. 

Consistency of model decoding. Thanks to its structural design, SiaTrans can be universally used for RGB-D and RGB saliency detection tasks. The metrics in Column F in Table~\ref{tab1} are obtained by SiaTrans by predicting the saliency of only RGB images in the test. The result is even better than some of the SOTA methods in Table~\ref{tab3}, sufficiently confirming that SiaTrans is qualified for RGB saliency detection tasks. 

\begin{figure*}[h!t]
\centering
\includegraphics[width=5in]{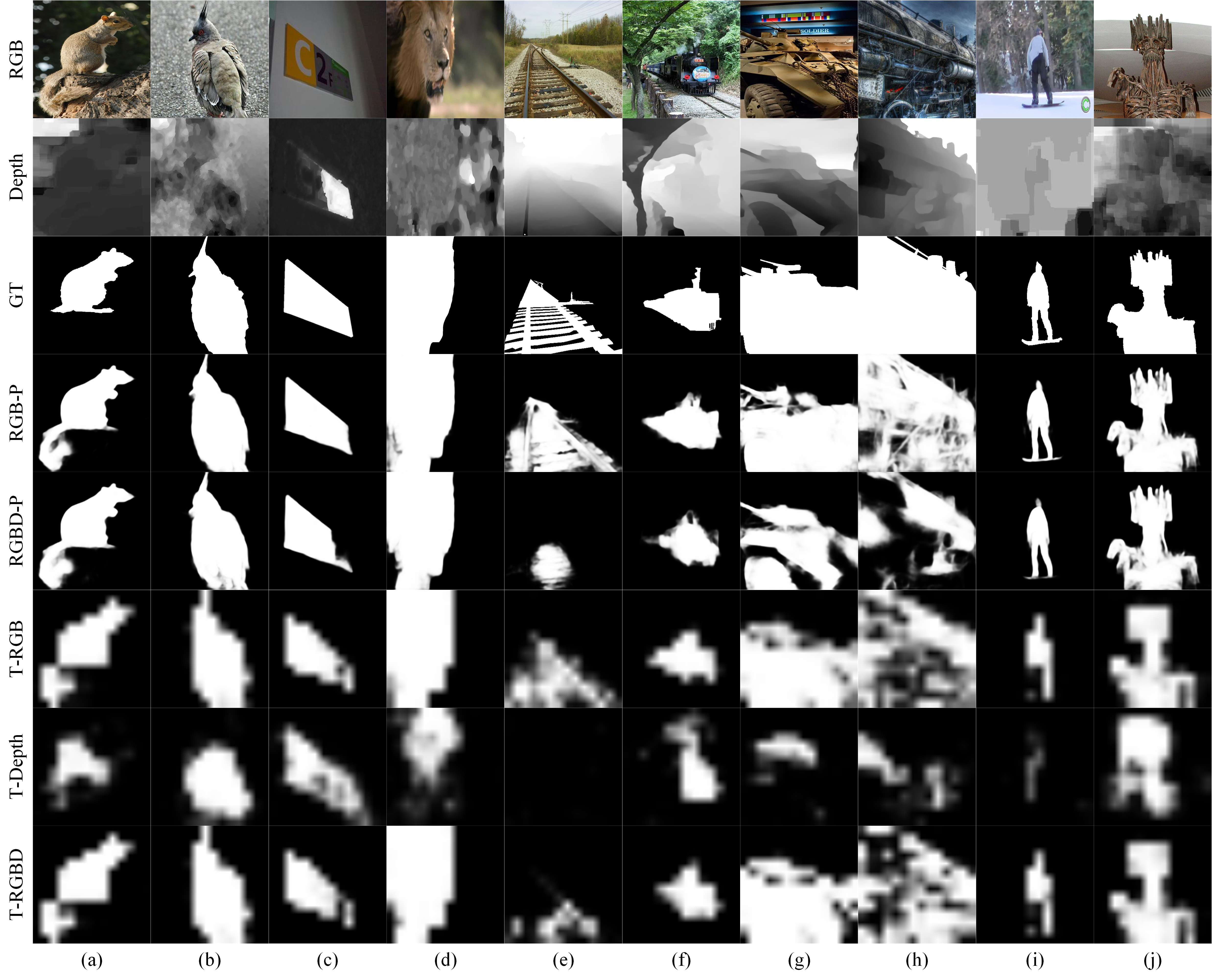}
\caption{Visualization examples of SiaTrans.}
\label{fig5}
\end{figure*}
Effectiveness of depth image classification. Fig.\ref{fig5} shows the image pairs with poor-quality depth image as judged by SiaTrans. As shown in Fig.\ref{fig5}, affected by the information interference of poor-quality depth images, the quality of the RGB-D saliency prediction map is lower than that of the RGB salience prediction map. SiaTrans has an ability to recognize poor-quality depth images and undertake the task of classifying depth images and RGB SOD. In Table~\ref{tab1}, the metrics in column G are improved and decreased compared with those in column E. There are three reasons to explain these changes. First, the proportion of poor-quality depth images is smaller in the RGB-D dataset. Second, our model is trained from the RGB-D dataset and is not sufficiently trained for single-model RGB. Third, as shown in Fig.\ref{fig5}(a), (b), the depth images are of poor quality, yet the RGB-D saliency prediction map is not much different from the RGB saliency prediction map.

\begin{table*}[h!t]
\center
\caption{Quantitative comparison of our proposed SiaTrans with other 7 SOTA RGB SOD methods on 6 benchmark datasets. “-R” means the ResNet50. Siatrans-RT means that it is trained by RGB dataset. Red and blue represent the best and second best results, respectively}
\resizebox{\textwidth}{!}{\begin{tabular}{cc|ccccccccc}
\hline
\multicolumn{2}{c|}{Metric}         & PoolNet & BASNet & ITSD-R                       & LDF-R                        & MINet                        & GateNet-R                    & VST                          & SiaTrans                     & SiaTran-RT                   \\ \hline
\multicolumn{2}{c|}{MACs(G)}        & 88.89   & 127.36 & 15.96                        & 15.51                        & 87.11                        & 162.13                       & 23.16                        & \textbf{10.91}               & \textbf{10.91}               \\
\multicolumn{2}{c|}{Params (M)}     & 68.26   & 87.06  & 26.47                        & 25.15                        & 162.38                       & 128.63                       & 44.48                        & \textbf{22.24}               & \textbf{22.24}               \\ \hline
                           & ${S_m} \uparrow$    & 0.852   & 0.837  & 0.861                        & 0.861                        & {\color[HTML]{000000} 0.856} & {\color[HTML]{3531FF} 0.863} & {\color[HTML]{FE0000} 0.873} & 0.860                        & {\color[HTML]{FE0000} 0.873} \\
                           & $F_\beta ^{max} \uparrow$ & 0.830   & 0.819  & 0.839                        & 0.839                        & 0.831                        & 0.836                        & 0.850                        & {\color[HTML]{3531FF} 0.851} & {\color[HTML]{FE0000} 0.864} \\
                           & $E_\phi ^{\max } \uparrow$ & 0.880   & 0.868  & 0.889                        & 0.888                        & 0.883                        & 0.886                        & 0.900                        & {\color[HTML]{3531FF} 0.910} & {\color[HTML]{FE0000} 0.915} \\
\multirow{-4}{*}{PASCAL-S} & $MAE \downarrow$   & 0.076   & 0.083  & 0.071                        & 0.067                        & 0.071                        & 0.071                        & 0.067                        & {\color[HTML]{3531FF} 0.063} & {\color[HTML]{FE0000} 0.060} \\ \hline
                           & ${S_m} \uparrow$    & 0.917   & 0.916  & 0.925                        & 0.925                        & 0.925                        & 0.924                        & {\color[HTML]{FE0000} 0.932} & 0.920                        & {\color[HTML]{3531FF} 0.930} \\
                           & $F_\beta ^{max} \uparrow$ & 0.929   & 0.931  & 0.939                        & 0.938                        & 0.938                        & 0.935                        & {\color[HTML]{FE0000} 0.944} & 0.932                        & {\color[HTML]{3531FF} 0.940} \\
                           & $E_\phi ^{\max } \uparrow$ & 0.948   & 0.951  & 0.959                        & 0.954                        & 0.957                        & 0.955                        & {\color[HTML]{FE0000} 0.964} & 0.959                        & {\color[HTML]{3531FF} 0.963} \\
\multirow{-4}{*}{ECSSD}    & $MAE \downarrow$   & 0.042   & 0.037  & {\color[HTML]{3531FF} 0.035} & {\color[HTML]{FE0000} 0.034} & {\color[HTML]{FE0000} 0.034} & 0.038                        & {\color[HTML]{FE0000} 0.034} & 0.037                        & {\color[HTML]{3531FF} 0.035} \\ \hline
                           & ${S_m} \uparrow$    & 0.832   & 0.836  & 0.840                        & 0.839                        & 0.833                        & 0.840                        & {\color[HTML]{FE0000} 0.850} & 0.828                        & {\color[HTML]{3531FF} 0.849} \\
                           & $F_\beta ^{max} \uparrow$ & 0.769   & 0.779  & 0.792                        & 0.782                        & 0.769                        & 0.782                        & {\color[HTML]{FE0000} 0.800} & 0.766                        & {\color[HTML]{3531FF} 0.797} \\
                           & $E_\phi ^{\max } \uparrow$ & 0.869   & 0.872  & {\color[HTML]{3531FF} 0.880} & 0.870                        & 0.869                        & 0.878                        & {\color[HTML]{FE0000} 0.888} & 0.877                        & {\color[HTML]{FE0000} 0.888} \\
\multirow{-4}{*}{DUT-O}    & $MAE \downarrow$   & {\color[HTML]{3531FF} 0.056}   & 0.057  & 0.061                        & {\color[HTML]{FE0000} 0.055} & {\color[HTML]{3531FF} 0.056}                        & {\color[HTML]{FE0000} 0.055} & 0.058                        & 0.061                        & {\color[HTML]{3531FF} 0.056}                        \\ \hline
                           & ${S_m} \uparrow$    & 0.797   & 0.772  & 0.809                        & 0.800                        & 0.805                        & 0.801                        & {\color[HTML]{FE0000} 0.820} & 0.788                        & {\color[HTML]{3531FF} 0.819}                        \\
                           & $F_\beta ^{max} \uparrow$ & 0.831   & 0.803  & {\color[HTML]{FE0000} 0.844} & 0.834                        & 0.836                        & 0.837                        & {\color[HTML]{3531FF} 0.843} & 0.812                        & {\color[HTML]{FE0000} 0.844} \\
                           & $E_\phi ^{\max } \uparrow$ & 0.858   & 0.827  & {\color[HTML]{3531FF} 0.869} & 0.862                        & 0.866                        & 0.866                        & {\color[HTML]{3531FF} 0.869} & 0.851                        & {\color[HTML]{FE0000} 0.875} \\
\multirow{-4}{*}{SOD}      & $MAE \downarrow$   & 0.105   & 0.112  & 0.093                        & 0.093                        & 0.093                        & 0.099                        & {\color[HTML]{FE0000} 0.087} & 0.104                        & {\color[HTML]{3531FF} 0.088} \\ \hline
\end{tabular}}
\label{tab2}
\end{table*}
Generalizability of the SiaTrans model. To verify the ability of our model to generalize RGB images, we test it on four RGB benchmark datasets (PASCAL-S \cite{34}, ECSSD \cite{35}, DUTOMRON \cite{36} and SOD \cite{37}) and compare it with seven RGB SOD models (PoolNet \cite{38}, BASNet \cite{39}, ITSD \cite{40}, LDF \cite{41}, MINet \cite{42}, CateNet-R \cite{43}, VST \cite{13}), as shown in Table~\ref{tab2}. It is noteworthy that, without modification, our SiaTrans model involves the least computation and the fewest parameters among these RGB SOD models. These figures prove the strong ability of our model to process RGB images. Here our SiaTrans model is trained from the RGB-D dataset with no separate training for RGB images. Comparison between the SiaTrans and SiaTrans-RT results indicates that, after filtering out poor-quality depth images, there is still room for improving SiaTrans’ training for single-modal RGB information. 

\subsection{Comparison with State-of-the-arts}
\begin{table*}[h!t]
\center
\caption{Quantitative comparison of our proposed model with 12 other SOTA RGB-D SOD methods on 9 benchmark datasets. Red and blue represent the best and second best results, respectively}
\resizebox{\textwidth}{!}{
\begin{tabular}{cc|ccccccccccccc}
\hline
\multicolumn{2}{c|}{Metric}         & ATST  & BBS-Net                      & CMMS                         & CoNet                        & DANet & HDFNet & JL-DCF                       & PGAR          & SSF   & UC-Net                       & TriTransNet                  & VST                          & Our                          \\
\multicolumn{2}{c|}{MACs(G)}        & 42.17 & 31.60                        & 134.77                       & 20.89                        & 66.25 & 91.77  & 211.06                       & 44.65         & 46.56 & 16.16                        & 223.82                       & 30.99                        & \textbf{10.91}               \\
\multicolumn{2}{c|}{Params (M)}     & 32.17 & 49.77                        & 92.02                        & 43.66                        & 26.68 & 44.15  & 143.52                       & \textbf{16.2} & 32.93 & 31.26                        & 139.54                       & 83.83                        & 22.24                        \\ \hline
                           & ${S_m} \uparrow$    & 0.885 & 0.921                        & 0.900                        & 0.896                        & 0.899 & 0.908  & 0.902                        & 0.909         & 0.899 & 0.897                        & 0.920                        & {\color[HTML]{3531FF} 0.922} & {\color[HTML]{FE0000} 0.923} \\
                           & $F_\beta ^{max} \uparrow$ & 0.893 & 0.919                        & 0.897                        & 0.893                        & 0.898 & 0.911  & 0.904                        & 0.907         & 0.896 & 0.895                        & {\color[HTML]{FE0000} 0.926} & 0.920                        & {\color[HTML]{3531FF} 0.921} \\
                           & $E_\phi ^{\max } \uparrow$ & 0.930 & 0.949                        & 0.936                        & 0.937                        & 0.935 & 0.944  & 0.944                        & 0.940         & 0.935 & 0.935                        & {\color[HTML]{3531FF} 0.955} & 0.951                        & {\color[HTML]{FE0000} 0.956} \\
\multirow{-4}{*}{NJUD}     & $MAE \downarrow$   & 0.047 & {\color[HTML]{3531FF} 0.035} & 0.044                        & 0.046                        & 0.046 & 0.039  & 0.041                        & 0.042         & 0.043 & 0.043                        & {\color[HTML]{FE0000} 0.030} & {\color[HTML]{3531FF} 0.035} & {\color[HTML]{3531FF} 0.035} \\ \hline
                           & ${S_m} \uparrow$    & 0.909 & {\color[HTML]{3531FF} 0.931} & 0.919                        & 0.912                        & 0.920 & 0.923  & 0.925                        & 0.917         & 0.915 & 0.920                        & 0.929                        & {\color[HTML]{FE0000} 0.932} & 0.929                        \\
                           & $F_\beta ^{max} \uparrow$ & 0.898 & 0.918                        & 0.904                        & 0.893                        & 0.909 & 0.917  & 0.918                        & 0.897         & 0.896 & 0.903                        & {\color[HTML]{FE0000} 0.924} & {\color[HTML]{3531FF} 0.920} & 0.918                        \\
                           & $E_\phi ^{\max } \uparrow$ & 0.951 & 0.961                        & 0.955                        & 0.948                        & 0.955 & 0.963  & 0.963                        & 0.950         & 0.953 & 0.956                        & {\color[HTML]{FE0000} 0.966} & 0.962                        & {\color[HTML]{3531FF} 0.964} \\
\multirow{-4}{*}{NLPR}     & $MAE \downarrow$   & 0.027 & 0.023                        & 0.028                        & 0.027                        & 0.027 & 0.023  & {\color[HTML]{3531FF} 0.022} & 0.027         & 0.027 & 0.025                        & {\color[HTML]{FE0000} 0.020} & 0.024                        & 0.024                        \\ \hline
                           & ${S_m} \uparrow$    & 0.916 & 0.882                        & 0.912                        & 0.923                        & 0.899 & 0.908  & 0.906                        & 0.899         & 0.915 & 0.871                        & 0.934                        & {\color[HTML]{FE0000} 0.943} & {\color[HTML]{3531FF} 0.940} \\
                           & $F_\beta ^{max} \uparrow$ & 0.928 & 0.870                        & 0.913                        & 0.932                        & 0.904 & 0.915  & 0.910                        & 0.898         & 0.923 & 0.864                        & 0.943                        & {\color[HTML]{FE0000} 0.948} & {\color[HTML]{3531FF} 0.944} \\
                           & $E_\phi ^{\max } \uparrow$ & 0.953 & 0.912                        & 0.940                        & 0.959                        & 0.939 & 0.945  & 0.941                        & 0.933         & 0.950 & 0.908                        & 0.964                        & {\color[HTML]{FE0000} 0.969} & {\color[HTML]{3531FF} 0.968} \\
\multirow{-4}{*}{DUTLF-D}  & $MAE \downarrow$   & 0.033 & 0.058                        & 0.036                        & 0.029                        & 0.042 & 0.041  & 0.042                        & 0.041         & 0.033 & 0.059                        & {\color[HTML]{3531FF} 0.025} & {\color[HTML]{FE0000} 0.024} & {\color[HTML]{3531FF} 0.025} \\ \hline
                           & ${S_m} \uparrow$    & 0.679 & 0.693                        & 0.699                        & 0.696                        & -     & 0.728  & 0.734                        & 0.656         & 0.595 & 0.713                        & 0.728                        & {\color[HTML]{FE0000} 0.759} & {\color[HTML]{3531FF} 0.753} \\
                           & $F_\beta ^{max} \uparrow$ & 0.673 & 0.680                        & 0.677                        & 0.693                        & -     & 0.717  & 0.727                        & 0.632         & 0.558 & 0.710                        & 0.735                        & {\color[HTML]{FE0000} 0.763} & {\color[HTML]{3531FF} 0.759} \\
                           & $E_\phi ^{\max } \uparrow$ & 0.758 & 0.763                        & 0.767                        & 0.782                        & -     & 0.804  & 0.805                        & 0.749         & 0.710 & 0.794                        & 0.802                        & {\color[HTML]{3531FF} 0.826} & {\color[HTML]{FE0000} 0.830} \\
\multirow{-4}{*}{ReDWeb-S} & $MAE \downarrow$   & 0.155 & 0.150                        & 0.143                        & 0.147                        & -     & 0.129  & 0.128                        & 0.161         & 0.189 & 0.130                        & {\color[HTML]{3531FF} 0.122} & {\color[HTML]{FE0000} 0.113} & {\color[HTML]{FE0000} 0.113} \\ \hline
                           & ${S_m} \uparrow$    & 0.896 & 0.908                        & 0.894                        & 0.905                        & 0.901 & 0.900  & 0.903                        & 0.894         & 0.837 & 0.903                        & 0.908                        & {\color[HTML]{3531FF} 0.913} & {\color[HTML]{FE0000} 0.914} \\
                           & $F_\beta ^{max} \uparrow$ & 0.901 & 0.903                        & 0.887                        & 0.901                        & 0.892 & 0.900  & 0.904                        & 0.880         & 0.840 & 0.899                        & {\color[HTML]{FE0000} 0.911} & 0.907                        & 0.907                        \\
                           & $E_\phi ^{\max } \uparrow$ & 0.942 & 0.942                        & 0.935                        & 0.947                        & 0.937 & 0.943  & 0.947                        & 0.929         & 0.912 & 0.944                        & {\color[HTML]{FE0000} 0.953} & {\color[HTML]{3531FF} 0.951} & {\color[HTML]{3531FF} 0.951} \\
\multirow{-4}{*}{STERE}    & $MAE \downarrow$   & 0.038 & 0.041                        & 0.045                        & {\color[HTML]{3531FF} 0.037} & 0.044 & 0.042  & 0.040                        & 0.045         & 0.065 & 0.039                        & {\color[HTML]{FE0000} 0.033} & 0.038                        & 0.038                        \\ \hline
                           & ${S_m} \uparrow$    & 0.850 & 0.863                        & 0.857                        & 0.851                        & 0.864 & 0.879  & 0.860                        & 0.832         & 0.790 & 0.865                        & 0.881                        & {\color[HTML]{3531FF} 0.889} & {\color[HTML]{FE0000} 0.897} \\
                           & $F_\beta ^{max} \uparrow$ & 0.853 & 0.843                        & 0.839                        & 0.837                        & 0.843 & 0.870  & 0.833                        & 0.798         & 0.762 & 0.855                        & 0.873                        & {\color[HTML]{3531FF} 0.876} & {\color[HTML]{FE0000} 0.889} \\
                           & $E_\phi ^{\max } \uparrow$ & 0.920 & 0.914                        & 0.900                        & 0.917                        & 0.914 & 0.925  & 0.902                        & 0.872         & 0.867 & 0.907                        & 0.934                        & {\color[HTML]{3531FF} 0.935} & {\color[HTML]{FE0000} 0.948} \\
\multirow{-4}{*}{SSD}      & $MAE \downarrow$   & 0.052 & 0.052                        & 0.053                        & 0.056                        & 0.050 & 0.046  & 0.053                        & 0.068         & 0.084 & 0.049                        & {\color[HTML]{3531FF} 0.041} & 0.045                        & {\color[HTML]{FE0000} 0.038} \\ \hline
                           & ${S_m} \uparrow$    & 0.849 & 0.879                        & 0.872                        & 0.860                        & 0.875 & 0.886  & 0.880                        & 0.838         & 0.799 & 0.875                        & 0.886                        & {\color[HTML]{FE0000} 0.904} & {\color[HTML]{3531FF} 0.899} \\
                           & $F_\beta ^{max} \uparrow$ & 0.861 & 0.884                        & 0.876                        & 0.873                        & 0.876 & 0.894  & 0.889                        & 0.827         & 0.786 & 0.879                        & 0.899                        & {\color[HTML]{FE0000} 0.915} & {\color[HTML]{3531FF} 0.913} \\
                           & $E_\phi ^{\max } \uparrow$ & 0.901 & 0.922                        & 0.911                        & 0.917                        & 0.918 & 0.930  & 0.925                        & 0.886         & 0.870 & 0.919                        & 0.930                        & {\color[HTML]{3531FF} 0.944} & {\color[HTML]{FE0000} 0.945} \\
\multirow{-4}{*}{SIP}      & $MAE \downarrow$   & 0.063 & 0.055                        & 0.058                        & 0.058                        & 0.055 & 0.048  & 0.049                        & 0.073         & 0.091 & 0.051                        & 0.043                        & {\color[HTML]{FE0000} 0.040} & {\color[HTML]{3531FF} 0.041} \\ \hline
                           & ${S_m} \uparrow$    & 0.917 & 0.934                        & 0.934                        & 0.914                        & 0.924 & 0.926  & 0.931                        & 0.886         & 0.904 & 0.934                        & -                            & {\color[HTML]{FE0000} 0.943} & {\color[HTML]{3531FF} 0.936} \\
                           & $F_\beta ^{max} \uparrow$ & 0.916 & 0.928                        & 0.928                        & 0.902                        & 0.914 & 0.921  & 0.923                        & 0.864         & 0.885 & 0.930                        & -                            & {\color[HTML]{FE0000} 0.940} & {\color[HTML]{3531FF} 0.932} \\
                           & $E_\phi ^{\max } \uparrow$ & 0.961 & 0.966                        & 0.969                        & 0.948                        & 0.966 & 0.970  & 0.968                        & 0.924         & 0.940 & {\color[HTML]{3531FF} 0.976} & -                            & {\color[HTML]{FE0000} 0.978} & 0.975                        \\
\multirow{-4}{*}{RGBD135}  & $MAE \downarrow$   & 0.022 & 0.021                        & {\color[HTML]{3531FF} 0.018} & 0.024                        & 0.023 & 0.022  & 0.021                        & 0.032         & 0.026 & 0.019                        & -                            & {\color[HTML]{FE0000} 0.017} & 0.020                        \\ \hline
                           & ${S_m} \uparrow$    & 0.845 & 0.835                        & 0.845                        & 0.848                        & 0.841 & 0.846  & 0.853                        & 0.808         & 0.851 & 0.856                        & 0.858                        & {\color[HTML]{FE0000} 0.882} & {\color[HTML]{3531FF} 0.871} \\
                           & $F_\beta ^{max} \uparrow$ & 0.858 & 0.828                        & 0.858                        & 0.852                        & 0.840 & 0.858  & 0.863                        & 0.794         & 0.863 & 0.860                        & 0.866                        & {\color[HTML]{FE0000} 0.889} & {\color[HTML]{3531FF} 0.876} \\
                           & $E_\phi ^{\max } \uparrow$ & 0.893 & 0.870                        & 0.886                        & 0.895                        & 0.874 & 0.889  & 0.894                        & 0.853         & 0.892 & 0.898                        & 0.898                        & {\color[HTML]{FE0000} 0.920} & {\color[HTML]{3531FF} 0.907} \\
\multirow{-4}{*}{LFSD}     & $MAE \downarrow$   & 0.078 & 0.092                        & 0.082                        & 0.076                        & 0.087 & 0.085  & 0.077                        & 0.099         & 0.074 & 0.074                        & 0.073                        & {\color[HTML]{FE0000} 0.061} & {\color[HTML]{3531FF} 0.069} \\ \hline
\end{tabular}
}
\label{tab3}
\end{table*}
\begin{figure*}[h!t]
\centering
\includegraphics[width=5in]{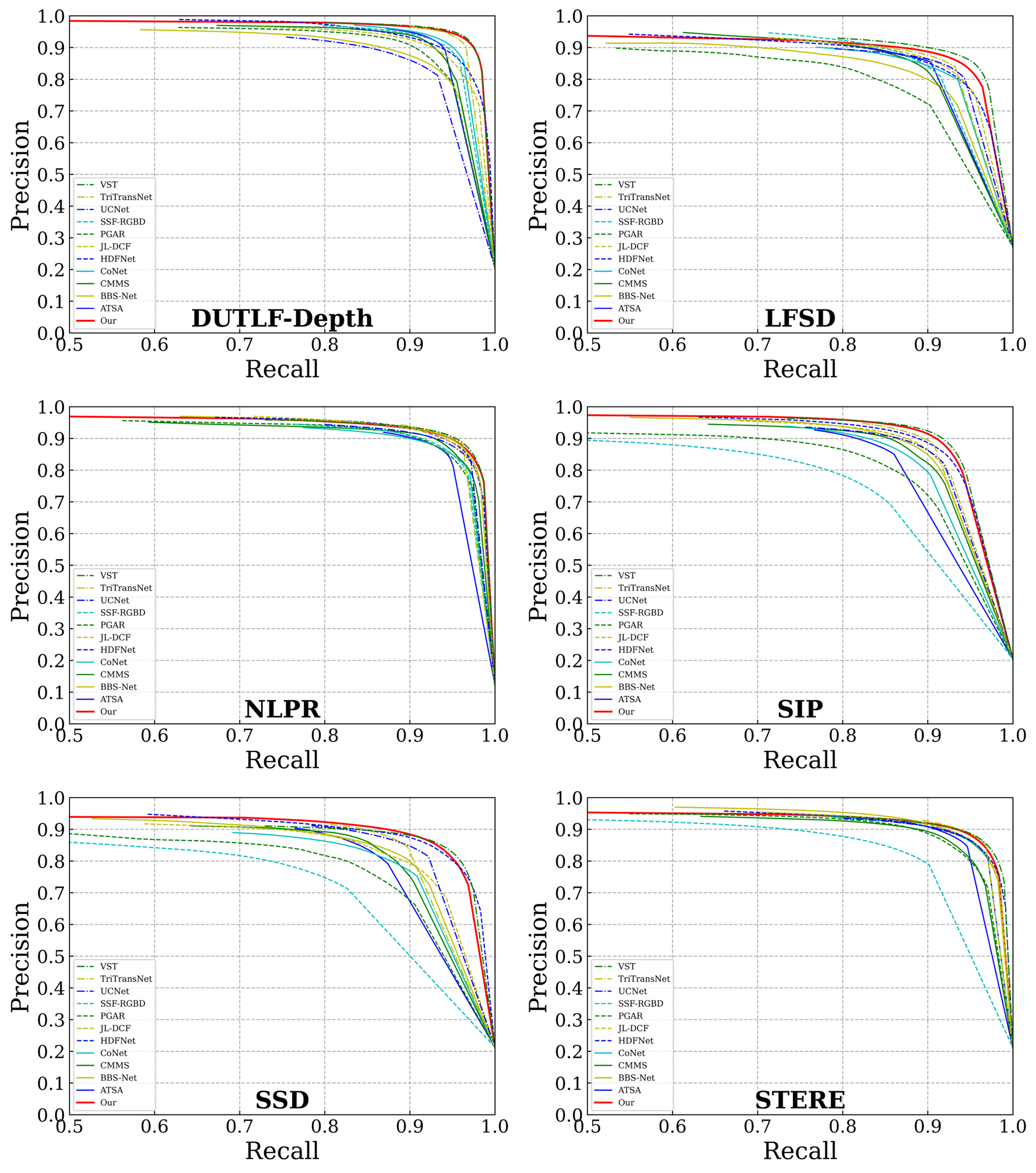}
\caption{The PR curves of proposed approach and other methods on six datasets.}
\label{fig6}
\end{figure*}
\begin{figure*}[h!t]
\centering
\includegraphics[width=5in]{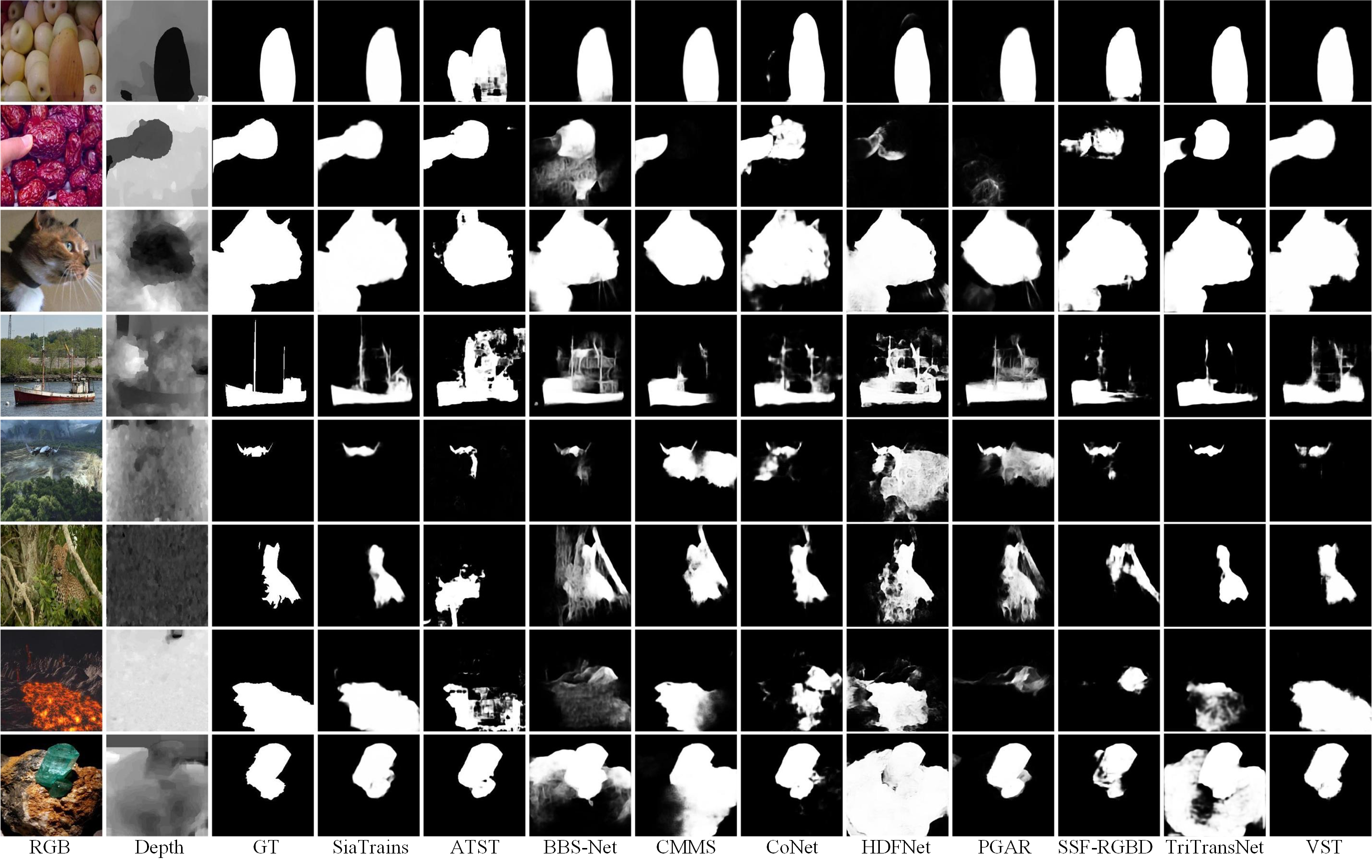}
\caption{Qualitative comparison with state-of-the-art RGB-D SOD methods.}
\label{fig7}
\end{figure*}
In this section, for RGB-D SOD tasks, we compare our model with 12 state-of-the-art methods, including ATST \cite{44}, BBS-Net \cite{9}, CMMS \cite{45}, CoNet \cite{46}, DANet \cite{12}, HDFNet \cite{10}, JL-DCF \cite{8}, PGAR \cite{47}, SSF-RGBD \cite{48}, UC-Net \cite{11}, TriTransNet \cite{14}, VST \cite{13}. Except TriTransNet and VST that are transformer based, all other methods are based on a convolutional network. To ensure that the comparison results are fair, the saliency maps for all these methods come from their authors or open-source work. Table~\ref{tab3} quantitatively compares the four evaluation metrics on the nine RGB-D benchmark datasets. Fig.~\ref{fig6} shows the PR curves. 

We can see from these results that our method compares favorably with the state-of-the-art methods in terms of the evaluation metrics, especially for the SSD dataset. The results in Table~\ref{tab2} and Table~\ref{tab3} indicate that, on the RGB and RGB-D datasets, our SiaTrans model outperforms many of the state-of-the-art RGB-D SOD and RGB SOD models, with a comparable number of parameters and the smallest MAC, proving the high effectiveness of our SiaTrans model. In Fig.~\ref{fig7}, we visually compare our SiaTrans with other state-of-the-art RGB-D SOD models. The results show that our SiaTrans can not only handle the impact of poor-quality depth images on saliency prediction, but also use depth information to predict saliency. Besides outputting prediction maps according to the depth classification result, it can also undertake RGB SOD tasks. Furthermore, our SiaTrans can also accurately detect salient objects in very challenging scenes, such as large salient objects, cluttering backgrounds, and low contrast ratios. 

\section{Conclusions}
We propose an RGB-D SOD framework called SiaTrans, which is the first to apply classification to RGB-D SOD tasks. SiaTrans not only shares a transformer network in the encoding stage for batch training to learn RGB and depth features, but also shares a network in the decoding stage to predict RGB and RGB-D saliency maps. Experiments indicate that it is feasible to train the model on depth image classification at the same time with saliency detection and satisfactory results can be achieved. This proves our observation that, through deep learning, computers are well able to recognize poor-quality depth images. Our model also maintains the least computation among non-lightweight models. SiaTrans outperforms recent state-of-the-art methods on nine evaluation datasets and has been verified by comprehensive ablation experiments. Besides RGB-D images, our model can also directly detect the saliency of separate RGB images with satisfactory performance. However, our SiaTrans has two shortcomings. One is that due to the limitation of datasets, our model is insufficiently trained for single-modal RGB in case of extremely poor-quality depth images. The other is that mistakes can be made when using MAE algorithm to produce depth image classification labels. In the future, we will continue to improve the model over these two problems. We also hope that this work will become a catalyst for many future RGB-SOD tasks.   
\subsection*{Acknowledgements}
This work was supported by SDUST Young Teachers Teaching Talent Training Plan(BJRC20180501); National Natural Science Foundation of China(Grant No. 61976125)

%
%
%
%
\bigskip
\bibliographystyle{splncs04}

\end{document}